\documentclass[letterpaper, 10 pt, conference]{ieeeconf}  

\IEEEoverridecommandlockouts                              

\usepackage{amsmath,amssymb,amsfonts}
\usepackage{algorithmic}
\usepackage{graphicx}
\usepackage{braket}
\usepackage{textcomp}
\usepackage{xcolor}
\usepackage{subcaption}
\usepackage{float}
\usepackage{multicol}
\usepackage{caption}
\usepackage{multirow, boldline, bbm, makecell}
\usepackage{hyperref}
\usepackage{colortbl,xcolor}
\usepackage[colorinlistoftodos]{todonotes}

\usepackage[normalem]{ulem} 

\definecolor{MyDarkBlue}{rgb}{0,0.08,1}
\definecolor{MyAqua}{rgb}{0,0.7,0.7}
\definecolor{MyDarkGreen}{rgb}{0.02,0.6,0.02}
\definecolor{MyDarkRed}{rgb}{0.8,0.02,0.02}
\definecolor{MyDarkOrange}{rgb}{0.40,0.2,0.02}
\definecolor{MyPurple}{RGB}{111,0,255}
\definecolor{MyRed}{rgb}{1.0,0.0,0.0}
\definecolor{MyGold}{rgb}{0.75,0.6,0.12}
\definecolor{MyDarkgray}{rgb}{0.66, 0.66, 0.66}
\definecolor{nicegreen}{rgb}{0.1, 0.6, 0.2}

\newcommand{\etal}{et al.~}

\title{\LARGE \bf
Taxim: An Example-based Simulation Model\\ for GelSight Tactile Sensors
}

\author{Zilin Si$^{1}$ and Wenzhen Yuan$^{1}$
\thanks{$^{1}$Zilin Si and Wenzhen Yuan are with the Robotics Institute, Carnegie Mellon University, 5000 Forbes Ave, Pittsburgh, PA, 15213, USA
        {\tt\small <zsi, wenzheny>@andrew.cmu.edu}}%
}

\begin{document}

\maketitle
\thispagestyle{empty}
\pagestyle{empty}

\begin{abstract}
Simulation is widely used in robotics for system verification and large-scale data collection. However, simulating sensors, including tactile sensors, has been a long-standing challenge. In this paper, we propose Taxim, a realistic and high-speed simulation model for a vision-based tactile sensor, GelSight~\cite{yuan2017gelsight}. A GelSight sensor uses a piece of soft elastomer as the medium of contact and embeds optical structures to capture the deformation of the elastomer, which infers the geometry and forces applied at the contact surface. We propose an example-based method for simulating GelSight: we simulate the optical response to the deformation with a polynomial look-up table. This table maps the contact geometries to pixel intensity sampled by the embedded camera. In order to simulate the surface markers' motion that is caused by the surface stretch of the elastomer, we apply the linear displacement relationship and the superposition principle. The simulation model is calibrated with less than 100 data points from a real sensor. The example-based approach enables the model to easily migrate to other GelSight sensors or its variations. To the best of our knowledge, our simulation framework is the first to incorporate \textit{marker motion field simulation} that derives from elastomer deformation together with the \textit{optical simulation}, creating a comprehensive and computationally efficient tactile simulation framework. Experiments reveal that our optical simulation has the lowest pixel-wise intensity errors compared to prior work and can run online with CPU computing. {Our code and supplementary materials are open-sourced at \href{https://github.com/CMURoboTouch/Taxim}{https://github.com/CMURoboTouch/Taxim}.}

\end{abstract}

\section{Introduction}

Simulation has been widely applied in robotics. It enables roboticists to quickly generate large amounts of realistic data, without costly equipment, manual labour, and the risk associated with real-world experiments.
With growing interest in robot simulation, well-developed simulation frameworks such as Gazebo~\cite{gazebo}, PyBullet \cite{coumans2016pybullet}, MuJoCo \cite{todorov2012mujoco}, Drake \cite{tedrake2019drake}, SOFA \cite{allard2007sofa}, NVIDIA Isaac Gym \cite{makoviychuk2021isaac} have been widely used in the robotics community. They can simulate dynamic rigid-body, soft-body, vision and laser sensors with varying levels of accuracy and speed. {However, none of them have integrated simulation of tactile sensing which form an irreplaceable part of robotic systems}. 

Recent advancements in vision-based tactile sensors, such as GelSight \cite{johnson2009retrographic}, \cite{yuan2017gelsight}, have made high-resolution tactile sensing available. 
\begin{figure}[t]
    \centering
    \includegraphics[width=0.48\textwidth]{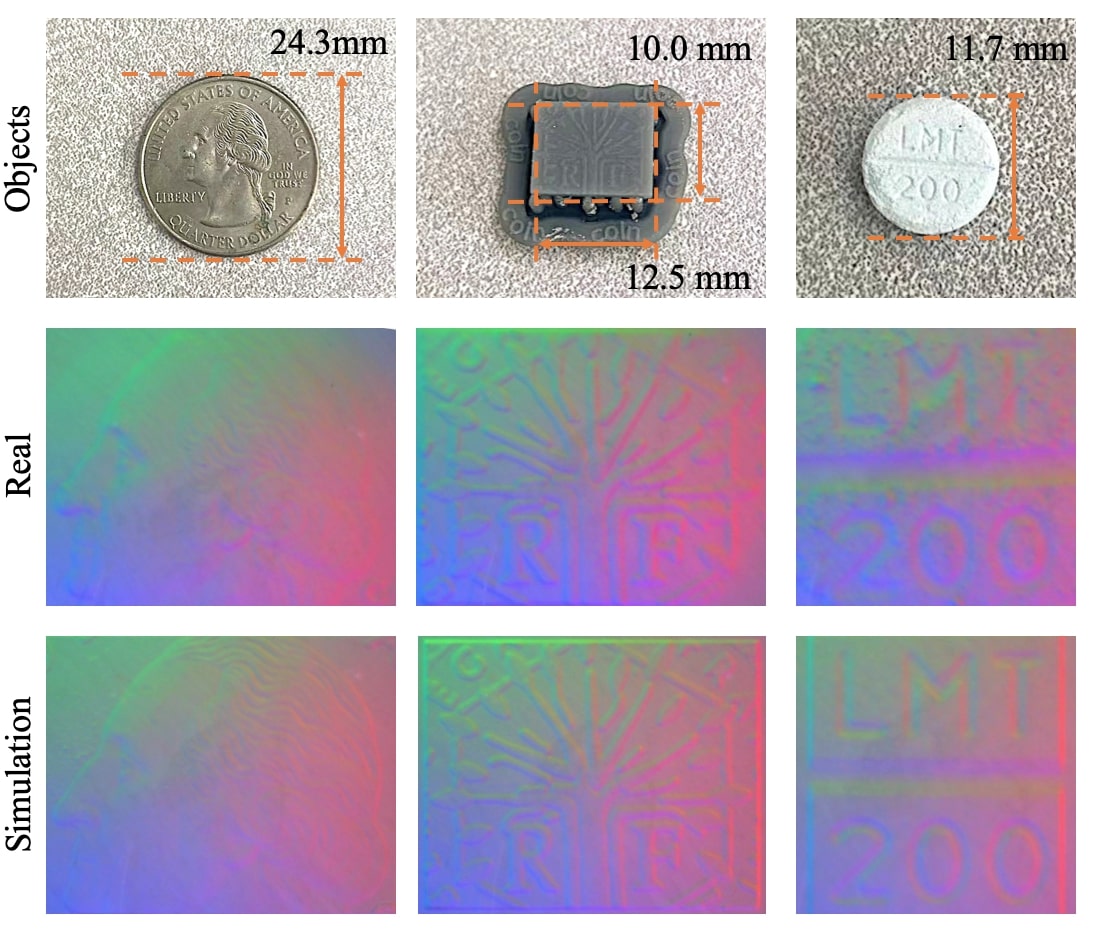}
    \caption{The GelSight outputs when it contacts objects with rich textures. With the ground-truth geometry~\cite{gkioulekas2015micron}, our simulation model generates images that are very similar to the real ones.}
    \label{fig:teaser}
\end{figure}
These sensors use a piece of soft elastomer, or \textit{gelpad}, as the contact medium for interacting with the environment. There is typically a printed marker array on the gelpad surface that moves as the surface stretches and is a good indicator of the contact forces and torques. The sensor utilizes optical components, including LEDs and an embedded camera to capture the illumination change caused by the change of light reflection on the galpad surface when the sensor contacts an external object, as shown in Fig~\ref{fig:optical}. 
Simulating those vision-based tactile sensors, which contains modeling both the mechanical response of the soft gelpad and the optical response to the deformation, is challenging. 
There have been previous studies on simulating different components of vision-based tactile sensors separately. For instance, Ding \etal\cite{ding2020sim} built a physics soft body simulation for the TacTip \cite{ward2018tactip} sensor to indicate pins' motion on the soft membrane; Agarwal \etal\cite{agarwal2021simulation} and Gomes \etal\cite{gomes2021generation} applied physics-based models for vision-based tactile optical simulation; whereas Wang \etal\cite{wang2020tacto} integrated the optical simulation of tactile sensors with the physics simulation engine PyBullet. However, the work mentioned above lack the ability to simulate the intrinsic noise of the real sensors. They also demand heavy computation while are difficult to generalize to new sensors.  


In this work, we propose \textit{Taxim}, an example-based tactile simulation model that combines \textit{optical simulation} and \textit{marker motion field simulation}. The method overcomes the constraints of the previous simulation models in that it is computationally lightweight and generates very similar signals to the real sensors in spite of the intrinsic noise of the sensors. The model contains two parts:
\begin{enumerate}
    \item A polynomial table mapping function to simulate the \textit{optical response} of a GelSight sensor by mapping geometries to pixel intensity in tactile images and an accumulation approach to simulate the shadow caused by the illumination.
    \item A model to simulate the \textit{marker motion field} using the linear displacement relationship and the superposition principle for the gelpad's elastic deformation.
\end{enumerate}

Taxim is calibrated with less than 100 contact examples,so that it can easily migrate to other vision-based tactile sensors with similar designs as GelSight. To the best of our knowledge, Taxim is the first model that simulates all functions of vision-based tactile sensors, including the optical response for geometry measurement and marker motion field for force/torque measurement. Our simulation model can be integrated into robot simulation engines to provide a useful tool for tacile-based manipulation research.  



\section{Related Work}

\subsection{Tactile Sensing Simulation}
The majority of tactile sensors use a soft medium for contact where the measurement of its deformation can indicate the contact information. Therefore the sensor simulation has been mostly focused on simulating the elastomer deformation. {Typically, elastic soft body simulation is modelled with finite element methods (FEM)~\cite{ma2019dense}, mass-spring model~\cite{hui2006interactive}, particles~\cite{wang2021elastic} or learning methods~\cite{casas2018learning}. To simulate tactile sensors that use soft medium, a traditional way is to build an approximation model of the soft bodies. Pezzementi \etal\cite{pezzementi2010characterization} and Moisio \etal\cite{moisio2012simulation} simulated the low dimensional tactile sensor signals with a point spread function model and a soft contact
model with a full friction description respectively. However, they are not applicable to the high-resolution vision-based tactile sensors such as GelSight. Narang \etal\cite{narang2020interpreting} used a finite element model to simulate the BioTac sensor~\cite{BioTac} and contact data in ANSYS. As an extension work, they simulated the deformation of BioTac in Issac gym~\cite{makoviychuk2021isaac} and projected this deformation to electronic signal with a generative learning framework in~\cite{narang2021sim}. Sferrazza \etal\cite{sferrazza2019ground, sferrazza2020learning} built a synthetic dataset with a finite element model of the vision-based tactile sensors, trained a network to predict the 3D contact force distribution in simulate and realized sim-to-real transfer. In this work, instead of using an accurate finite element model with high-computing cost, we approximate the deformation of the soft medium on the GelSight sensor with pyramid Gaussian kernels which is efficient and also gives acceptable accuracy.}

\subsection{Optical Simulation for Tactile Sensors}
For vision-based tactile sensors like GelSight, optical simulation is essential as it is used to measure geometries of the contacted objects. To simulate the optical system of GelSight, Gomes \etal\cite{gomes2021generation} and Hogan \etal\cite{hogan2021seeing} used Phong's model to simulate the reflection and illumination. Agarwal \etal\cite{agarwal2021simulation} applied ray tracing to simulate the light paths within the sensor that form tactile images. Wang \etal\cite{wang2020tacto} presented TACTO, an open-source simulator using pyrender to simulate DIGIT~\cite{lambeta2020digit} sensors and bridged it to a physics simulator PyBullet. Compared to those physics-based methods, our method is data-driven, so that it is computationally efficient and can better simulate the intrinsic noise of the real sensors.

\subsection{Marker Motion Field Simulation for Tactile Sensors}
The movement of marker array on GelSight or other vision-based tactile sensors is caused by the planar stretch of the elastomer surface. They also make the key component of many vision-based tactile sensors such as TacTip~\cite{ward2018tactip}. In manipulation tasks such as slip detection~\cite{yuan2015measurement} and grasping stability prediction~\cite{calandra2017feeling}, the marker motion serves as an essential feature. For TacTip, Ding \etal\cite{ding2020sim} simulated the dynamics of its soft membrane in Unity so as to extracted markers' motion. They evaluated the simulation on sim-to-real robot tasks. {Church \etal\cite{church2021optical} simulated the depth maps to represent the contact geometries instead of optical tactile images. They also used a  Generative Adversarial Networks (GANs) to realize real-to-sim translation for TacTip sensors.} Unlike the above work, we explicitly simulate the marker motion field by using FEM offline and applying the superposition principle~\cite{cotin1999real} online. Our method does not require extensive training data but only a gelpad FEM model, and it approximates the marker motion field well with high accuracy and low computation cost.

\section{Methods}
\subsection{Overview}
\begin{figure}[t]
\centering
\includegraphics[width=0.48\textwidth]{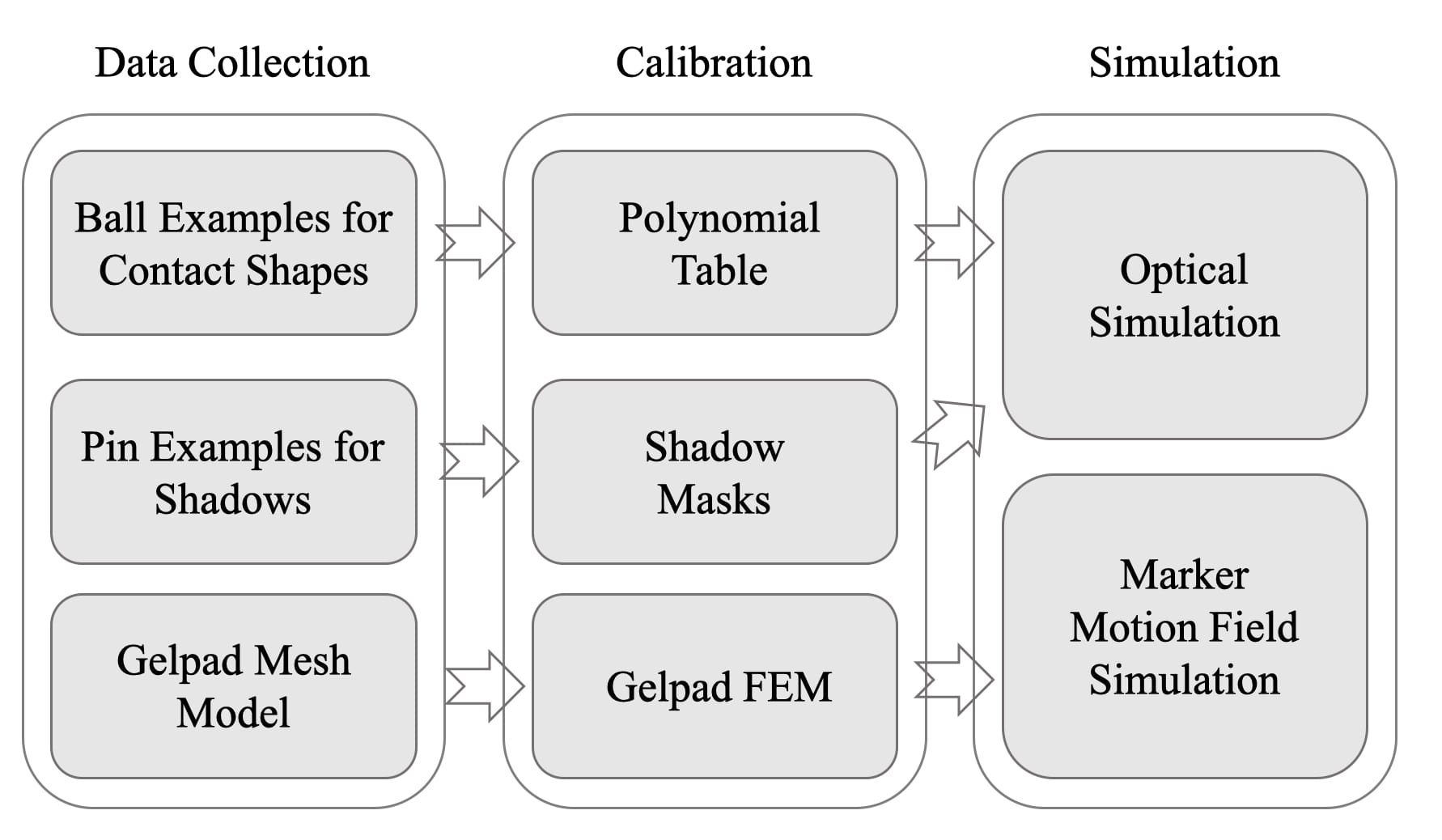}
\caption{The pipeline of our proposed example-based simulation model.} 
\label{fig:pipeline}
\end{figure}

We construct and employ our simulation models for both optical response and marker motion of GelSight sensors. To simulate the sensor's optical response, we build a polynomial table to map the contact geometries to the image intensities and collect shadow masks to attach the shadow. Then we apply the superposition principle based on the loading displacement of each finite unit to simulate the markers' motion. Their combination replicates the contacting of objects on the tactile sensor. For both parts, we calibrate our simulator with examples from a real sensor. We show the pipeline for building and applying the simulation model in Fig.~\ref{fig:pipeline}.

\subsection{Optical Simulation}
\label{sec:optical}
\begin{figure}[t]
\centering
\includegraphics[width=0.48\textwidth]{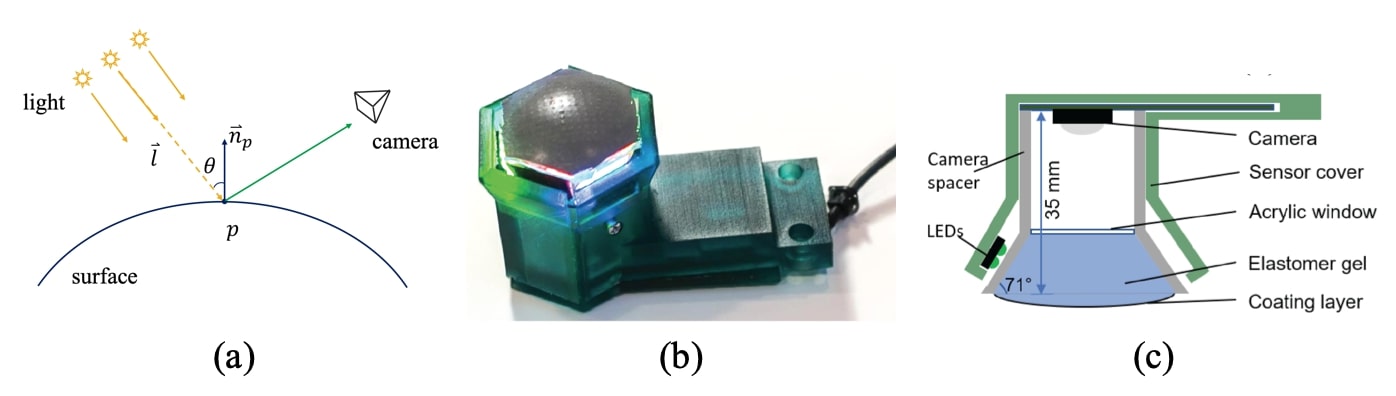}
\caption{(a) Demo of the photometric stereo method: for a surface point $p$ under the light $l$, the reflected light intensity captured by the camera is determined by the surface reflectance and the surface normal $\vec{n}_p$ (b) The GelSight sensor~\cite{dong2017improved} we aim to simulate and (c) its schematic diagrams of the optical structure.} 
\label{fig:optical}
\end{figure}
We simulate the optical response of the GelSight sensor as a result of the contact geometry with a model using the examples-based photometric stereo~\cite{hertzmann2005example} method. Photometric stereo uses the linear reflection function to derive the illumination of the object with the light sources and shape of the illuminated surface. Example-based photometric stereo does not require prior knowledge of lights sources but instead uses the imaging of the reference objects. 
We use a lookup table as the baseline and a polynomial table as our proposed method to map the contact shapes to image intensities.


\paragraph{Lookup Table Mapping}
The gelpad has a homogeneous diffuse internal surface which makes the reflection function spatial-invariant. The linear reflection function used by photometric stereo is formalized as $I_{p} = \rho \mathbf{n_p} \sum \mathbf{l}$, where at a point $p$, the observed light intensity $I_p$ caused by reflection is a product of the albedo $\rho$, the surface normal $\mathbf{n_p}$ and the light direction and intensity $\mathbf{l}$, as shown in Fig.~\ref{fig:optical}~(a). The previous equation implies that the reflected light intensity $I$ and the surface normal $\textbf{n}$ are linearly correlated. Alternatively, instead of solving the equation from the given lighting conditions, an intensity-shape lookup table can be built as follows
\begin{equation}
    I = \sum_{l} a^l \textbf{n}
\end{equation}
where $a^l$ is the coefficient of the light $l$, and can derived from a calibration process similar to~\cite{johnson2009retrographic, johnson2011microgeometry}. Here, we define the lights of the same color--any of red, green or blue--as one light source, even though they are contributed by multiple LEDs.  

\paragraph{Polynomial Table Mapping}
The linear lookup table works well with the point lights which are far from the object, whose directions are parallel and intensities are uniform for all the points on the illuminated surface. However, the LEDs in the GelSight are close to the sensor surface so that the emitted light is not strictly parallel and uniform. To compensate for the complicated lighting conditions, we introduce a non-linear model for the reflection as proposed in~\cite{angelopoulou2014uncalibrated}. The reflection function can then be rewritten as
\begin{equation}
    I = \sum_{l} f^{l}_{\mathbf{n}}(x, y)
\end{equation}
where $\mathbf{n}$ is the normal vector representative of the surface shape and $(x,y)$ is the 2D location on the image plane.
From experiments, we found that in practice a second order polynomial function is sufficient to approximate the non-linearity. Thus, the non-linear function is represented as:
\begin{equation}
    f^{l}_{\mathbf{n}}(x,y) = \mathbf{w}^l_\mathbf{n}\mathbf{b}
\end{equation}
where $\mathbf{w}^l_\mathbf{n}$ is a $6\times 1$ vector that represents the parameters to model the polynomial table, and $\mathbf{b}=[x^2, y^2, xy, x, y, 1]^{T}$.

\begin{figure}[t]
    \includegraphics[width=0.49\textwidth]{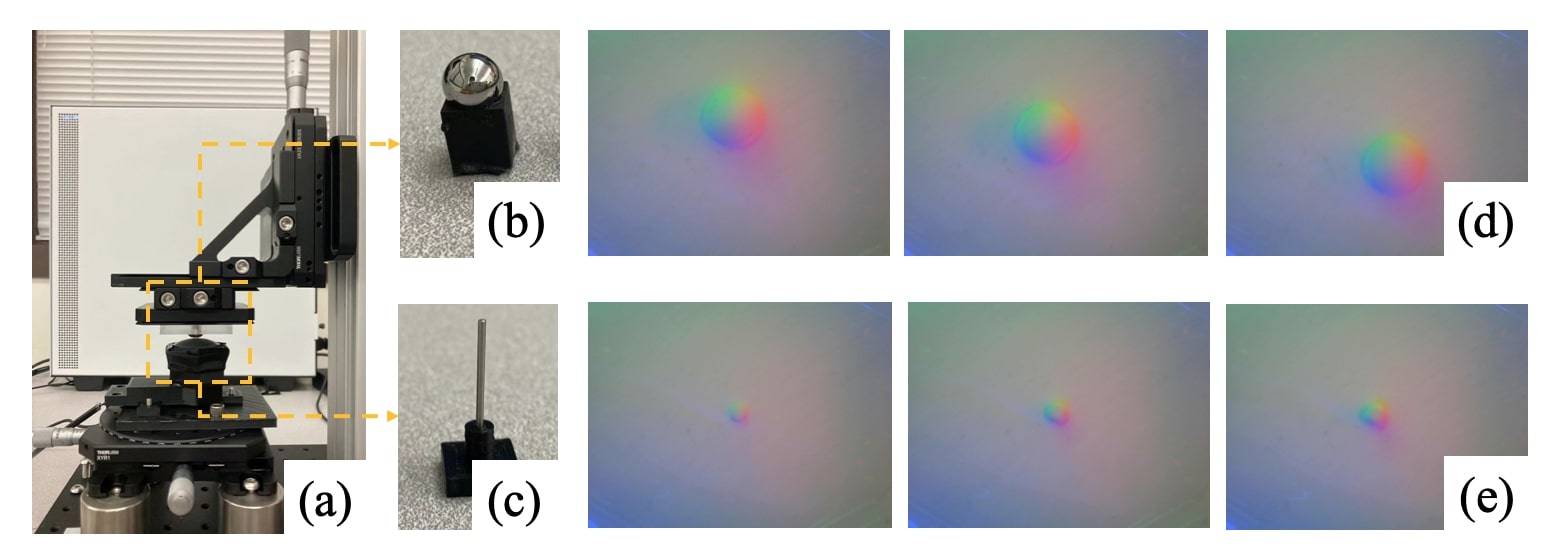}
    \caption{Data collection setup (a, b, c) and data examples (d, e) to build the optical simulation model. The GelSight is placed on an XYR optical stage and an indenter with a certain shape object is mounted on a vertical linear stage for precisely indenting on the GelSight. We calibrate the polynomial table with less than 100 data points using a spherical indenter and collect shadow masks with 10 data points using a pin indenter.} 
    \label{fig:calib_ball}
\end{figure}

\paragraph{Calibration}
Calibration entails fitting the parameters in the polynomial table from real data. Since these parameters vary for different sensors, this process has to be done per sensor. During calibration, we press a small ball with a known radius over the surface and manually locate contact areas in the tactile images as shown in Fig~\ref{fig:calib_ball} (b) and (d). The surface normal at each point in the circular area can be easily calculated based on the ball's geometry. We discretize the 3D surface normal vectors to a $125 \times 125$ table with the magnitude and direction of the surface normal as the two dimensions. The parameters in polynomial table can then be solved via least squares with the set of intensity-shape-location pairs $(I_p, \mathbf{n}_p, x_p, y_p)$ from collected data. We fill invalid values in the table by interpolation. 

\paragraph{Simulation}
We simulate the visual outputs in three steps: collision detection, deformation approximation, and optical simulation. A collision is detected when an object comes in contact with the gelpad. From this contact, the local shape, represented as a height map, is constructed from the object's shape in the contact area, and gelpad's shape in non-contact area. Additionally, we need to simulate the soft body deformation from the height map. An approximation of soft body simulation is applied with pyramid Gaussian kernels. The shape in contact area is kept unchanged to maintain the fine textures and the boundaries between the contact and non-contact areas are smoothed using pyramid Gaussian kernels from large to small. From the height map, the normal vector for each point can be extracted and mapped to an intensity value with the calibrated polynomial table to synthesize the tactile images.

\subsection{Shadow Simulation}
\label{sec:shadow}
\begin{figure}[t]
    \centering
    \includegraphics[width=0.4\textwidth]{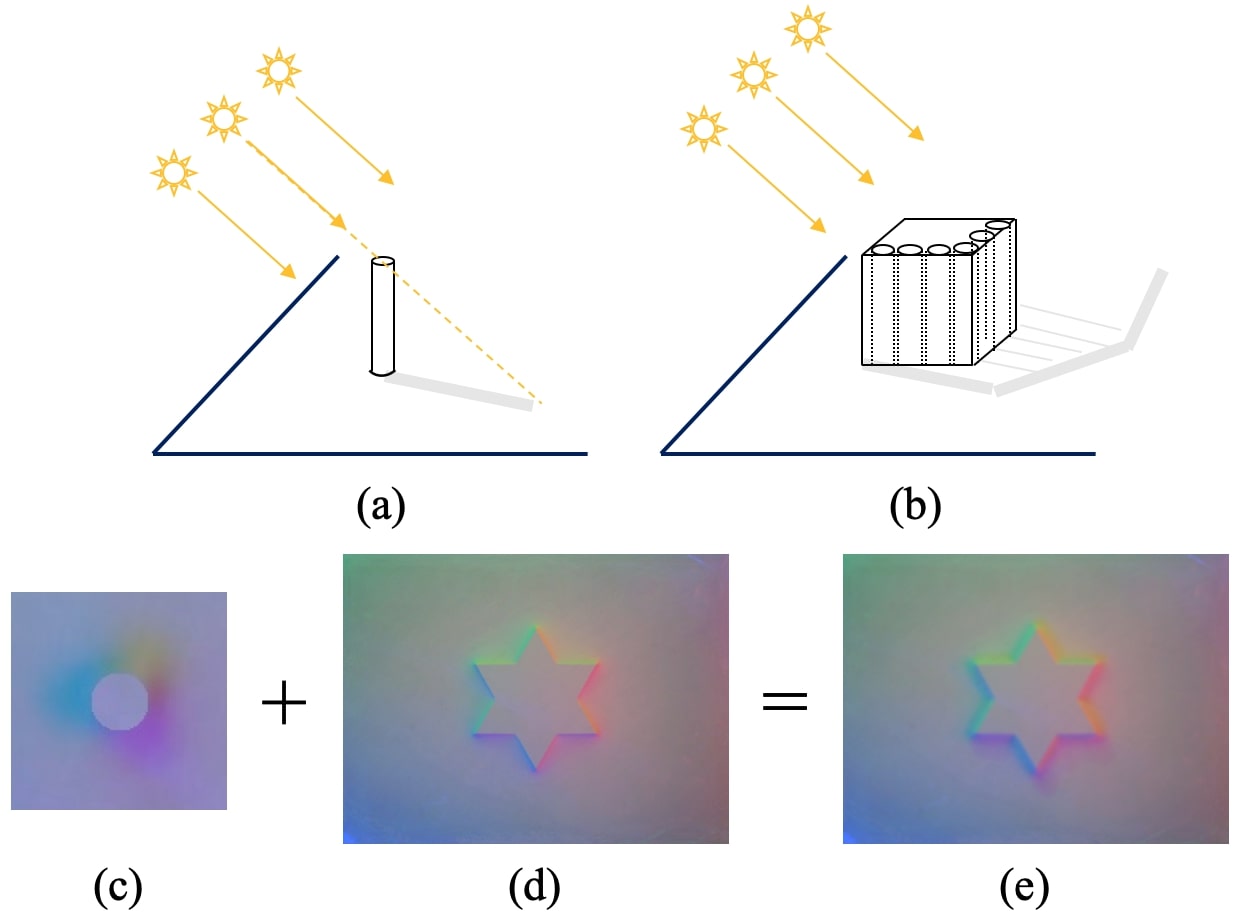}
    \caption{Shadow synthesis. (a) A unit shadow case observed under the lighting. (b) We approximate the object as the composition of unit pin case and then attach the shadow caused by each pin. (c), (d) and (e): We collect a set of shadow masks and synthesize the shadow around the contact area.} 
    \label{fig:shadow}
    \vspace{-2mm}
\end{figure}
Other than the illumination change that is modeled with photometric stereo method, the shadow is another factor causing the change of the pixel intensity in the tactile images. According to the design of the sensor, the shadows are caused by three groups of LED lights: red, green and blue lights. We simulate the shadow from those light sources respectively.
We simplify shadow casting by collecting the ``unit" shadow case, and then simulate the shadow by accumulating the shadows caused by each geometrical ``unit". Since each light beam is traveling independently in the space, without considering inter-reflection, the shadow cast by them can be linearly accumulated.

A ``unit" shadow is the shadow cast by a standing pin, as shown in Fig.~\ref{fig:shadow} (a). For objects with different geometries, we consider them as the accumulation of ``unit" shadows placed side by side with different heights, as illustrated in Fig.~\ref{fig:shadow} (b). Therefore, given the tactile images with shadow cast by indenting a pin normally onto the gelpad with different depths as shown in Fig.~\ref{fig:calib_ball} (c) and (e), we extract a set of shadow masks on three dominant directions caused by three light sources. For a general case, the shadow mask is attached for all three color channels and all points within the contact area if the neighbors are lower than that point.

\subsection{Marker Motion Field Simulation}
\label{sec:motion}
\begin{figure*}[t]
\includegraphics[width=0.98\textwidth]{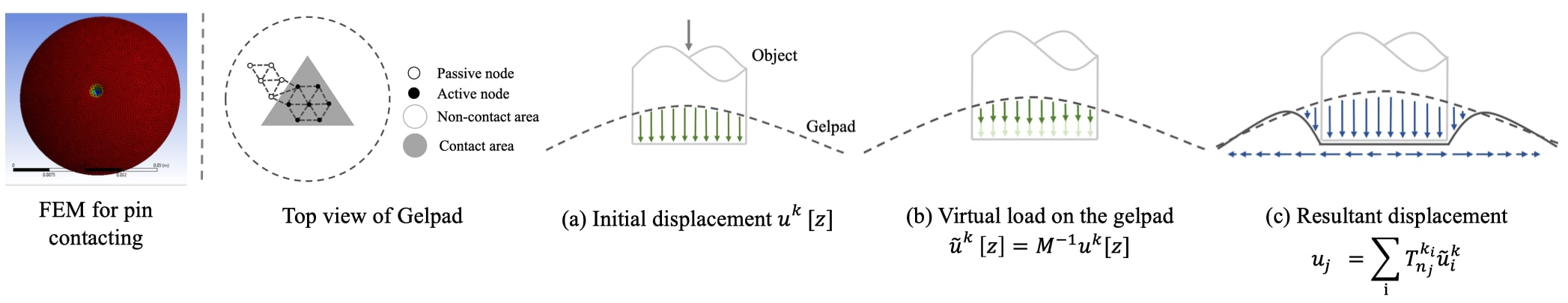}
\caption{Elastic deformation calibration and simulation for the gelpad. We calibrate the deformation of gelpad under a unit pin with 0.5 mm diameter indenting in ANSYS to get the dense nodal displacement results (left). In simulation, we compose each node's displacement from elastic deformation in three steps: (a) initial displacement boundary conditions on active nodes, (b) active nodes' virtual displacements, and (c) resultant nodal displacements for both active nodes in contact area and passive nodes in non-contact area.} 
\label{fig:elastic}
\vspace{-2mm}
\end{figure*}
We simulate the markers' motion on the gelpad surface caused by the deformation of the soft gelpad from contacting. In this work, we consider the deformation under normal and shear loads. We employ the linear displacement relationship and superposition principle~\cite{cotin1999real} to compose the deformation of the surface with loads on each finite unit of the contact surface. Although the markers are sparsely spread on the surface, we mesh the surface with dense nodes in simulation, track each node's motion and then locate the markers' motions. With the dense solution, the method can be applied to markers at any locations. Nodes in the gelpad surface mesh are classified into two categories: \textit{active nodes} who come in contact with the object and are applied external forces and constrained by internal elastic forces; \textit{passive nodes} who are in non-contacted area and only constrained by the internal elastic forces.

{The linear displacement relationship assumes that any two nodes can influence each other in a linear way. By considering two nodes $n_i$ and $n_j$ with displacement in 3D as $\mathbf{u_i}$ and $\mathbf{u_j}$, the $n_j$ can be passively influenced by $n_i$ as $\mathbf{u_j} = T^{n_i}_{n_j} \mathbf{u_i}$, where $T^{n_i}_{n_j}$ is a $3 \times 3$ tensor representing the mutual influence. The superposition principle states that a node $n_i$'s displacement $\mathbf{u_i}$ is an aggregation of all active nodes' influence to it. Assume we have active nodes $K = \{k_1, k_2, k_3, ..., k_m\}$, where $\mathbf{u^k_i}$ represent active nodes' initial displacement under external loads. Under the linear displacement relationship and superposition principle, any node $n_j$'s displacement $\mathbf{u_j}$ can be composed as
\begin{equation}
    \mathbf{u_j} = \sum^{m}_{i=1} T_{n_j}^{k_i} \mathbf{u^k_i}
\end{equation}}

{However, before applying the superposition principle by using the active nodes' initial displacement $\mathbf{u^k_i}$, we need to amend them to virtual displacements under the virtual loads because they not only are constrained by external loads but also influence each other. For instance, if all the active nodes' displacements are initialized such that they only move along the z direction \textit{i.e.} $\mathbf{u^k} = [0,0,dz]$, the following equation holds:}
\begin{equation}
    \mathbf{u^k_j}[z] = \sum^{m}_{i=1} T^{k_i}_{k_j}[3,3]\mathbf{\tilde{u}^k_i}[z]
\end{equation}
where $\mathbf{u^k_j}$ is the initialized displacement, and $\mathbf{u^k_j}[z]$ is its component along z-direction; $\mathbf{\tilde{u}^k_i}$ is the virtual displacement under the virtual load; $T^{k_i}_{k_j}[3,3]$ is the last element in the tensor $T^{k_i}_{k_j}$. Therefore, it is able to solve virtual displacements for active nodes by stacking all the equations as:
\begin{equation}
\begin{gathered}
    \mathbf{u^k}[z] = M_z\mathbf{\tilde{u}^k}[z] \\
    \left[\begin{matrix}
   \mathbf{u^k_1}[z] \\
   \mathbf{u^k_2}[z] \\
   ... \\
   \mathbf{u^k_m}[z]
   \end{matrix} \right] =
   \left[\begin{matrix}
   T^{1}_{1}[3,3] & T^{1}_{2}[3,3] & ... & T^{1}_{m}[3,3] \\
   T^{2}_{1}[3,3] & T^{2}_{2}[3,3] & ... & T^{2}_{m}[3,3] \\
   ... \\
   T^{m}_{1}[3,3] & T^{m}_{2}[3,3] & ... & T^{m}_{m}[3,3] \\
   \end{matrix} \right] 
   \left[\begin{matrix}
   \mathbf{\tilde{u}^k_1}[z] \\
   \mathbf{\tilde{u}^k_2}[z] \\
   ... \\
   \mathbf{\tilde{u}^k_m}[z]
   \end{matrix} \right]
\end{gathered}
\end{equation}
Then $\mathbf{\tilde{u}^k}[z]$ is solved by matrix inversion as $\mathbf{\tilde{u}^k}[z] = M_z^{-1}\mathbf{u^k}[z]$.

The $x$, $y$ components of the active nodes' displacement can be amended using the same approach, but with $T[1,1]$ or $T[2,2]$ for the $x$ or $y$ directions respectively. Later, we apply the superposition principle to get the final resultant displacements for all nodes with
\begin{equation}
    \mathbf{u_j} = \sum_{i} T_{n_j}^{k_i} \mathbf{\tilde{u}^k_i}
\end{equation}

\paragraph{Calibration}
The tensor $T^{k_i}_{n_j}$ depends on the gelpad's physical properties, and therefore can be measured in advance. The markers on the real gelpad are sparsely distributed which can not be used to generate dense meshes. Instead, we calibrate the arbitrary $T^{k_i}_{n_j}$ in a Finite Element Method (FEM) software ANSYS. In ANSYS, we generate the dense mesh of the gelpad and measure the deformation when there is a load on a unit node, as shown in Fig.~\ref{fig:elastic} (left). We then use the measurement of the deformation to calibrate $T$. Since the markers are not printed on the top surface of the gelpad, we extract the second layer's mesh which is 0.5mm below the top surface from the simulated model as reference. To fully calibrate the $3\times 3$ tensors, we simulate an active node's motion in z-direction only, a combination of z-direction and x-direction and a combination of z-direction and y-direction. Then we solve all the tensors $T^{k_i}_{n_j}$ using least squares from these three sets of unit case.


\paragraph{Simulation}We employ the marker motion simulation in three steps, by: 1) applying the initial displacements on the active nodes under the external loads, 2) getting active nodes' virtual displacements with the superposition principle, and then 3) calculating the resultant displacements at each node using the superposition principle with virtual displacements of active nodes. This process is demonstrated in Fig.~\ref{fig:elastic} (a), (b), (c).

\section{Experiments}
We perform a set of experiments to evaluate the similarity between the simulated tactile data and that from real sensors.
\subsection{Experiment Setup and Data Collection}
To collect well-controlled contact data with a real GelSight sensor, We set up an optical platform, as  shown in Fig.~\ref{fig:calib_ball} (a). The GelSight is placed on a XYR stage, and an indenter is mounted on a vertical linear stage positioned above the GelSight. We manually control the contact location and depth by adjusting the stages. The XYR stage enables horizontal movement and the vertical stage adjusts the indenting depth. Both are with 0.01mm precision. We use a dome-shaped gelpad for both the real sensor and the simulated sensor.

\begin{figure}[t]
\includegraphics[width=0.49\textwidth]{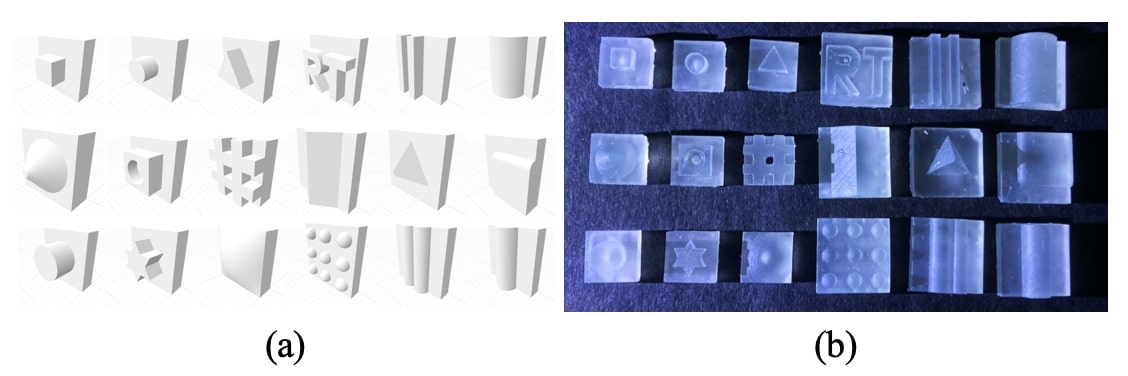}
\caption{Dataset of objects designed in Solidworks (a) and 3D printed (b) for contact experiments. The objects are of different shapes. Their base sizes are either 10mm $\times$ 10mm or 15mm $\times$ 15mm.}
\label{fig:dataset}
\vspace{-2mm}
\end{figure}

We evaluate our simulation using objects with different shapes and textures. The objects are designed in Solidworks~\cite{solidworks}, output as mesh files for simulation (Fig.~\ref{fig:dataset} (a)) and 3D printed for collecting data from the real sensor for comparison (Fig.~\ref{fig:dataset} (b)).

\subsection{Optical Simulation}
To calibrate the optical simulation model, we collect 50 data points on different locations of gelpad surface with a 4mm-diameter spherical indenter, as shown in Fig.~\ref{fig:calib_ball} (b), (d); to calibrate the shadow simulation model, we collect 10 data points of different pressing depths with a 1mm-diameter pin indenter, as shown in Fig.~\ref{fig:calib_ball} (c), (e). {The calibration process is simple and easy to conduct manually without precise control of contact locations which can be accomplished within 1 hour. And it can be used till any components of the sensor are replaced or the sensor is broken.}


\begin{figure}[t]
\includegraphics[width=0.49\textwidth]{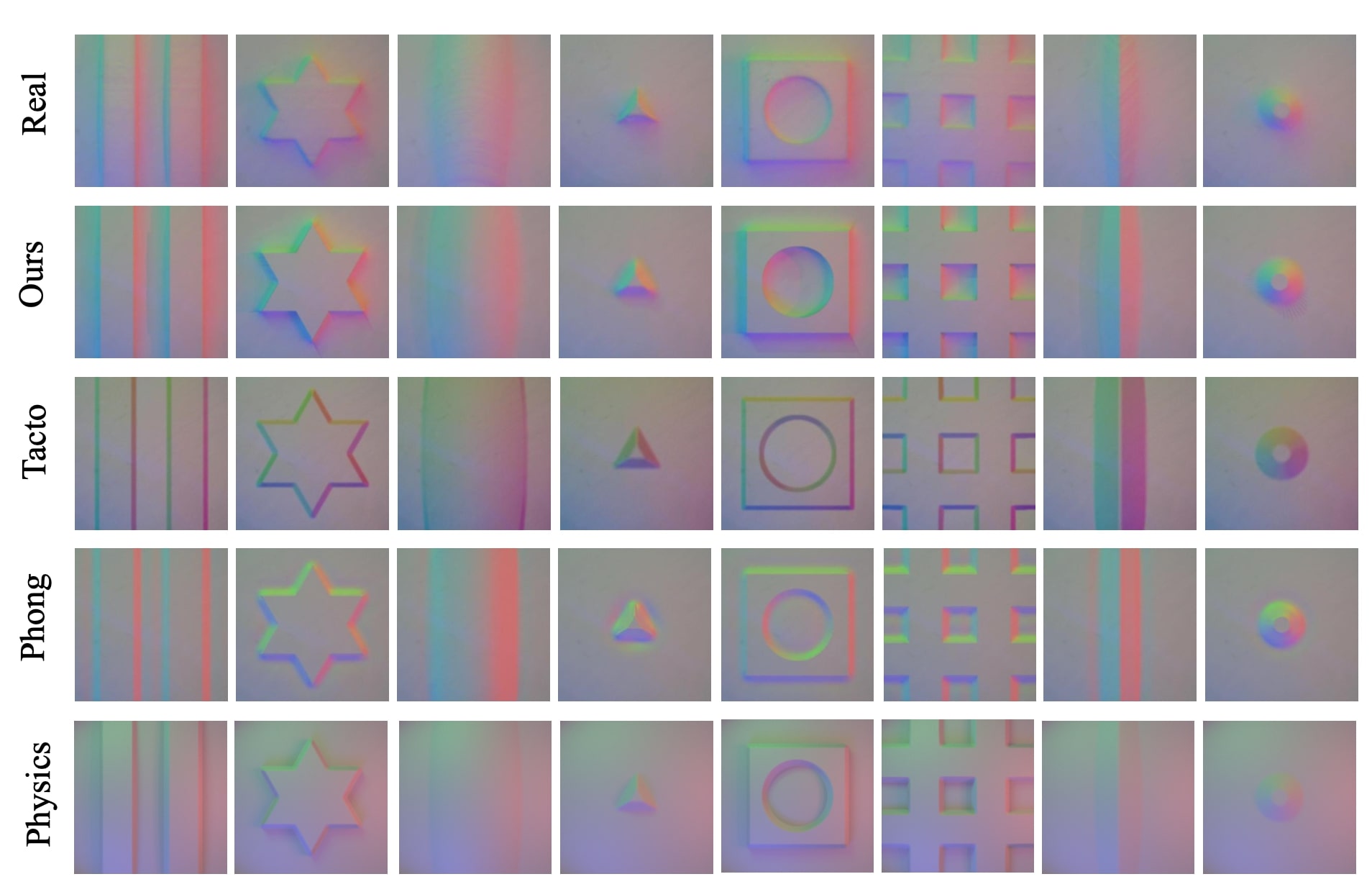}
\caption{Optical simulation comparison among our method, TACTO~\cite{wang2020tacto}, Phong~\cite{gomes2021generation} and physics~\cite{agarwal2021simulation} with the real data.}
\label{fig:all_results}
\end{figure}

We simulate the tactile images on the aforementioned dataset and compare our method with three other methods: the physics-based model~\cite{agarwal2021simulation}, TACTO~\cite{wang2020tacto} and Phong's model~\cite{gomes2021generation} as shown in Fig.~\ref{fig:all_results}. We evaluate our method by comparing the simulated images with the real images in pixel-wise level, against the three methods mentioned above on four metrics: mean absolute error (L1), mean squared error (MSE), structural index similarity (SSIM) and peak singal-to-noise ratio (PSNR). The simulated images are cropped to size $400 \times 400$ around the indenting area to eliminate the background's effect. Also, due to the precision of the operation with the real sensor, the ground truth tactile images are not well aligned with the simulated images. So we manually align the images using GIMP~\cite{gimp}. The quantitative results are summarized in the Table.~\ref{table:optical}. From the table, our method outperforms all the other methods.

\begin{table}[]
\centering
\begin{tabular}{l|l l l l}
\Xhline{1pt}
        & \textbf{L1} $\downarrow$            & \textbf{MSE} $\downarrow$             & \textbf{SSIM} $\uparrow$           & \textbf{PSNR} $\uparrow$            \\ \hline
Tacto~\cite{wang2020tacto}   & 10.861         & 215.861         & 0.808          & 25.495          \\ 
Phong's~\cite{gomes2021generation} & 8.163          & 123.249         & 0.832          & 27.763          \\ 
Physics~\cite{agarwal2021simulation} & 7.409         & 90.623         & 0.759          & 28.687          \\ 
\textbf{Ours} & \textbf{5.565} & \textbf{58.358} & \textbf{0.882} & \textbf{30.974} \\ \Xhline{1pt}
\end{tabular}
\caption{Image similarity metrics between simulation and real data for optical simulation. We compare our method with methods from Physics-based model~\cite{agarwal2021simulation}, Tacto~\cite{wang2020tacto} and Phong's model~\cite{gomes2021generation} on L1, MSE, SSIM and PSNR metrics. Our method performs the best on all the metrics.}
\label{table:optical}
\vspace{-2mm}
\end{table}

\begin{figure}[t]
\includegraphics[width=0.49\textwidth]{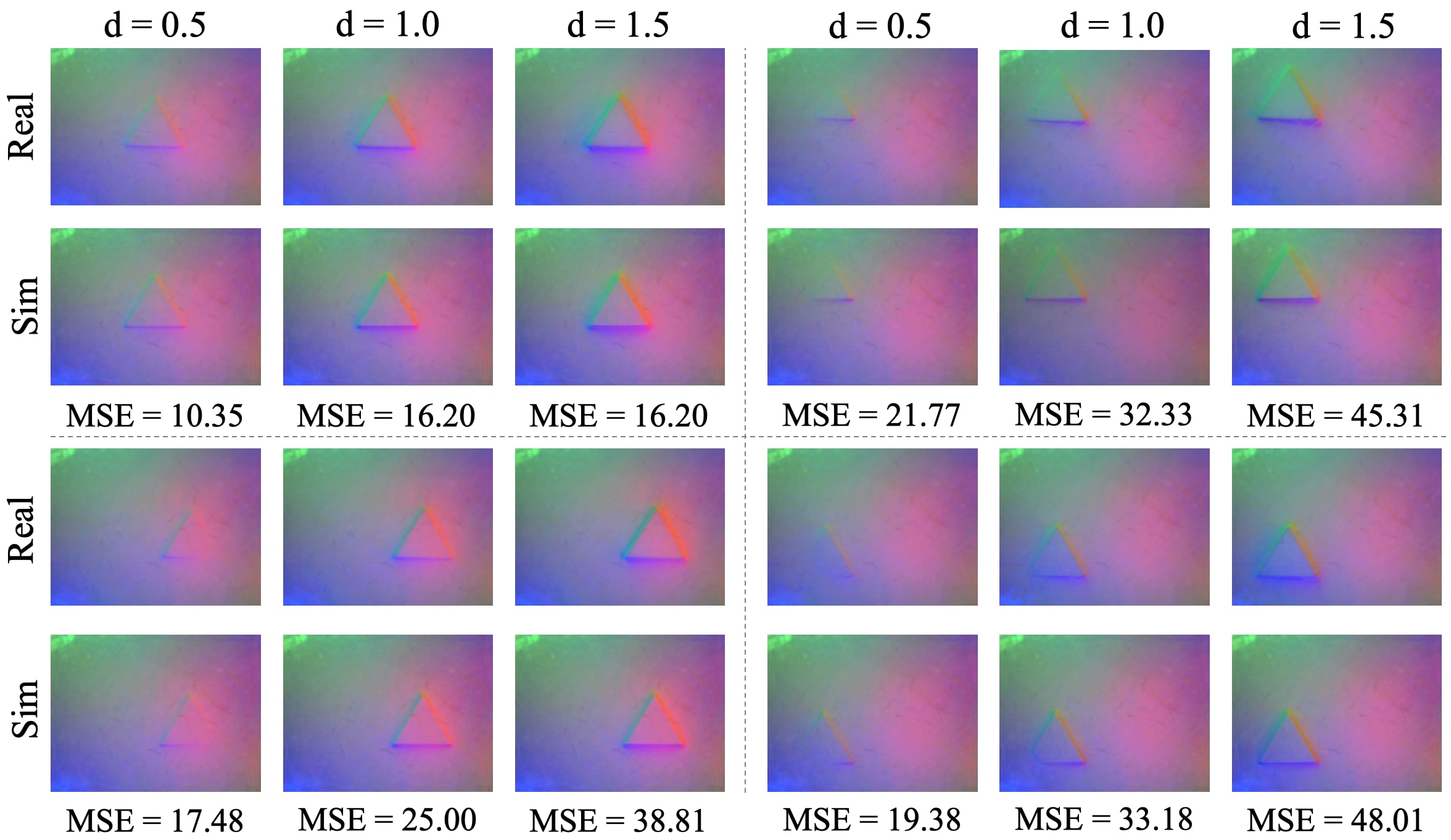}
\caption{{Optical simulation results with different indentation depths and locations. The locations differ over the gelpad surface while the depths differ as 0.5mm, 1.0mm, 1.5mm. The MSE error is shown below each pair.}}
\label{fig:var}
\end{figure}

{\textbf{Different indentation depths and locations} Our optical simulation model works well for different indentation depths and locations. One example is shown in Fig.~\ref{fig:var}. From MSE errors, we can see errors increase when the indentation become farther from the center and deeper.}

\textbf{Fine texture simulation}
Our model can simulate the contact cases with fine-textured objects, as shown in Fig.~\ref{fig:teaser}.

\begin{figure}[t]
\includegraphics[width=0.48\textwidth]{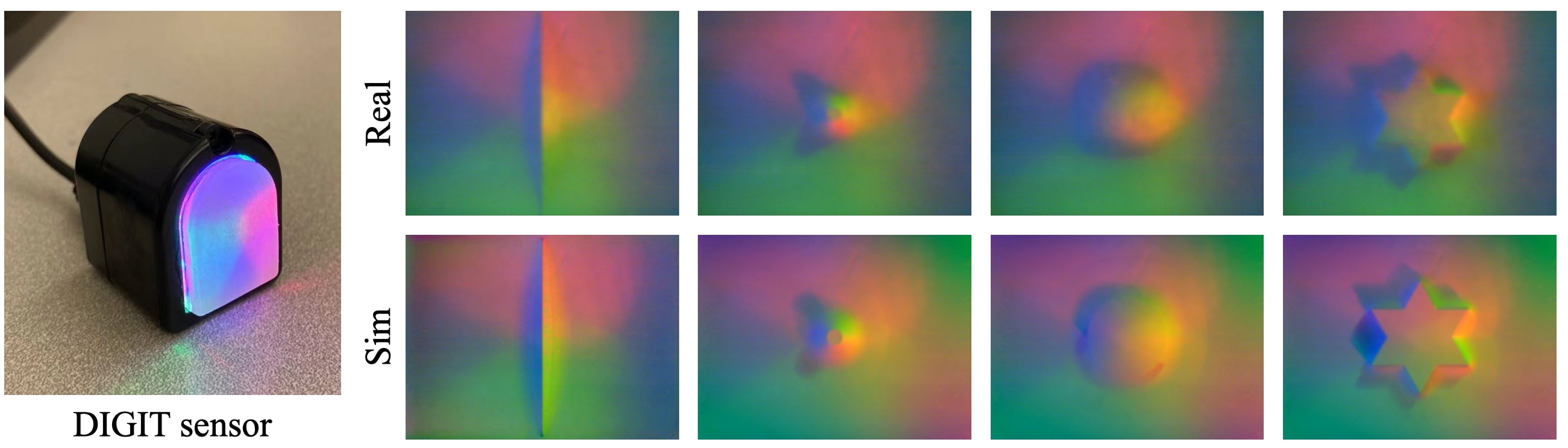}
\caption{{Optical simulation results (right) for a DIGIT sensor (left).}}
\label{fig:digit}
\vspace{-2mm}
\end{figure}

\begin{figure}[t]
\includegraphics[width=0.48\textwidth]{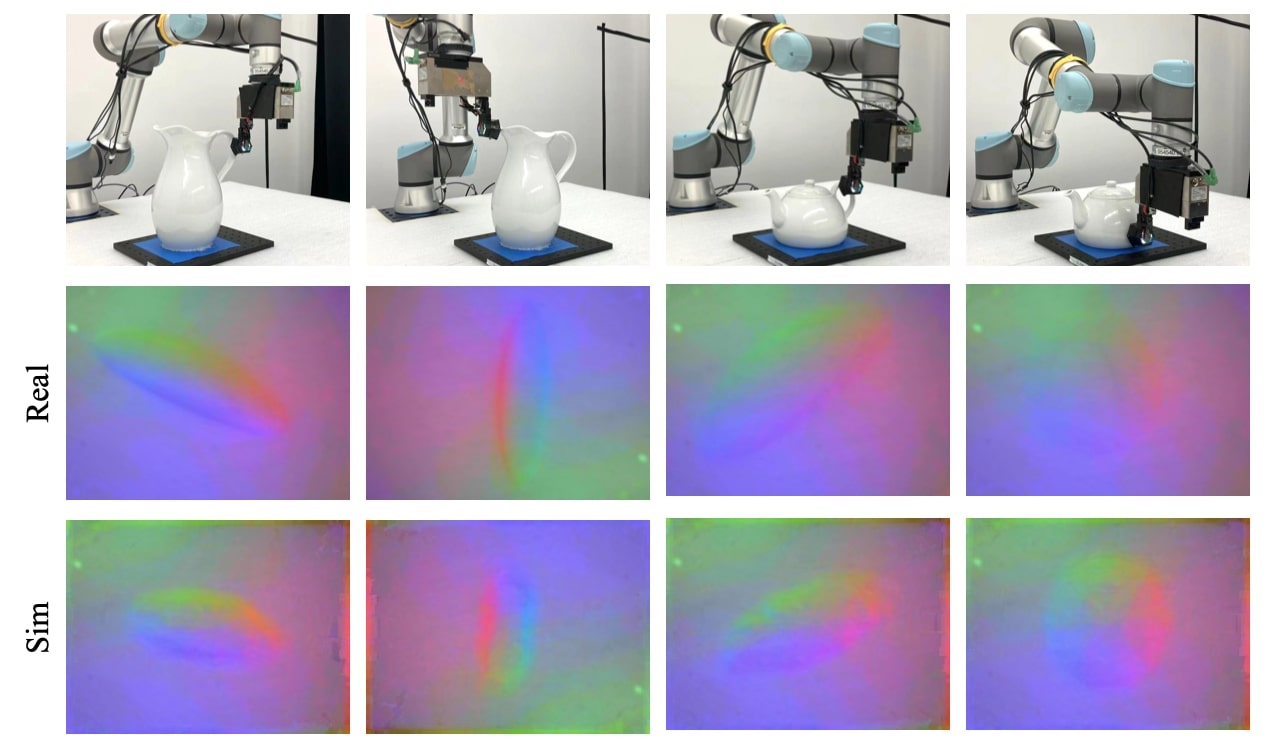}
\caption{{Optical simulation results for objects from the Google Scan dataset~\cite{googleScannedObjects}. We touch the objects with a GelSight mounted on a robot arm given certain contact locations. Most artifacts in the simulated images come from the coarse mesh files of the objects.}} 
\label{fig:real_obj}
\end{figure}

\begin{figure*}[t]
\includegraphics[width=0.98\textwidth]{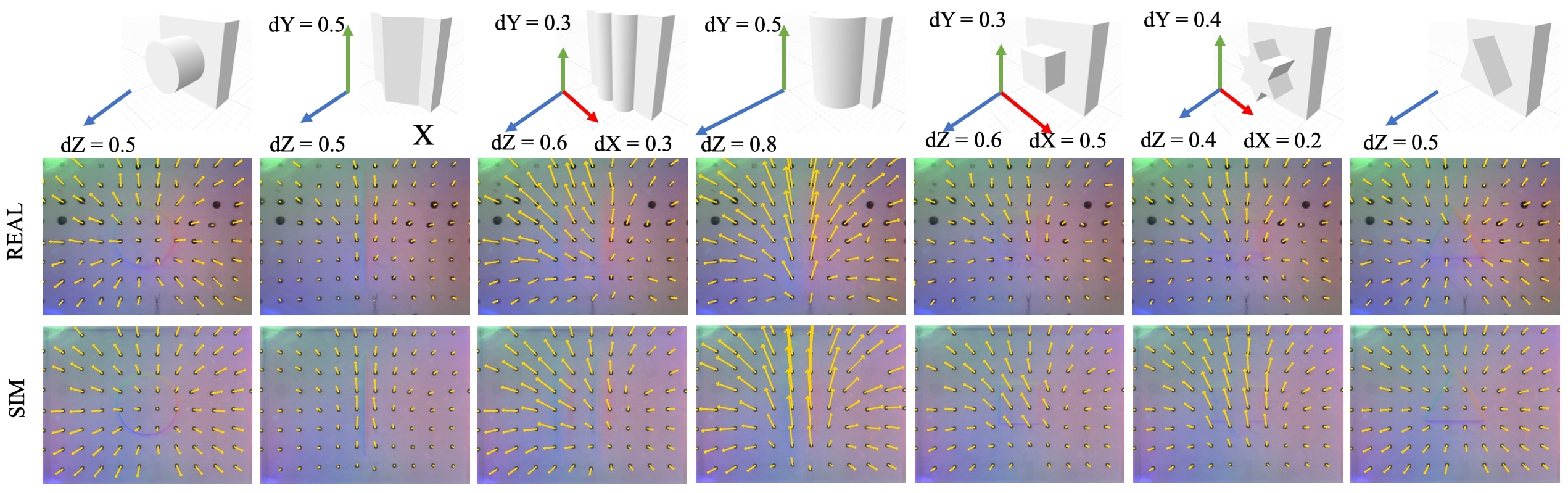}
\caption{Marker motion field simulation with optical simulation results. We visualize the marker motions (scaled up by 20 for better visualization)  on the dataset under different normal displacements and shear displacements.}
\label{fig:all_markers_1}
\vspace{-2mm}
\end{figure*}

{\textbf{Simulation on various sensors and objects}
Note that tactile images look different in Fig.~\ref{fig:all_results}, Fig.~\ref{fig:var} Fig.~\ref{fig:real_obj} and Fig.~\ref{fig:all_markers_1}. This is because we use 4 different GelSight sensors and manufactures lead to the difference. However, our model works well on all of them. We also apply our model on a DIGIT sensor~\cite{lambeta2020digit} and the results are shown in Fig.~\ref{fig:digit}. In addition, We test our model on various objects from the Google Scan dataset~\cite{googleScannedObjects} and some results are shown in Fig.~\ref{fig:real_obj}.}

\textbf{Speed test}
We test all the simulation techniques, mentioned above, on a AMD Ryzen Threadripper 2950X 16-Core Processor CPU.
We input height maps with the size 480 $\times$ 640, and output the simulated tactile images of the dataset. We then record the average running time of all the methods, as shown in the Table~\ref{table:speed}. {Our method is the most computationally lightweight on CPU and achieves the real-time data transferring speed from real sensors.} However, Tacto~\cite{wang2020tacto} and Physics model~\cite{agarwal2021simulation} can be largely accelerated on GPUs but not considered here for evaluation. Our method can be potentially optimized for GPU computation as well and we will work on that for the next step.
\begin{table}[H]
\centering
\begin{tabular}{l|l l l l l }
\Xhline{1pt}
\textbf{Speed}      & \begin{tabular}[c]{@{}l@{}}\textbf{Ours w/o} \\\textbf{shadows}\end{tabular} & \begin{tabular}[c]{@{}l@{}}\textbf{Ours w/} \\\textbf{shadows}\end{tabular} & \begin{tabular}[c]{@{}l@{}}\textbf{Physics} \\~\cite{agarwal2021simulation}\end{tabular} & \begin{tabular}[c]{@{}l@{}}\textbf{Tacto} \\~\cite{wang2020tacto}\end{tabular} & \begin{tabular}[c]{@{}l@{}}\textbf{Phong's} \\~\cite{gomes2021generation}\end{tabular} \\ \hline
Frequency (fps) & 18.1                  & 9.6                  & 0.1      & 1.9   & 3.8     \\ \Xhline{1pt}
\end{tabular}
\caption{Speed test for optical simulation on CPU. We compare our method with the Physics-based model~\cite{agarwal2021simulation}, Tacto~\cite{wang2020tacto} and Phong's model~\cite{gomes2021generation}. Our method runs with the fastest speed.}
\label{table:speed}
\vspace{-5mm}
\end{table}

\begin{figure}[t]
    \centering
    \includegraphics[width=0.49\textwidth]{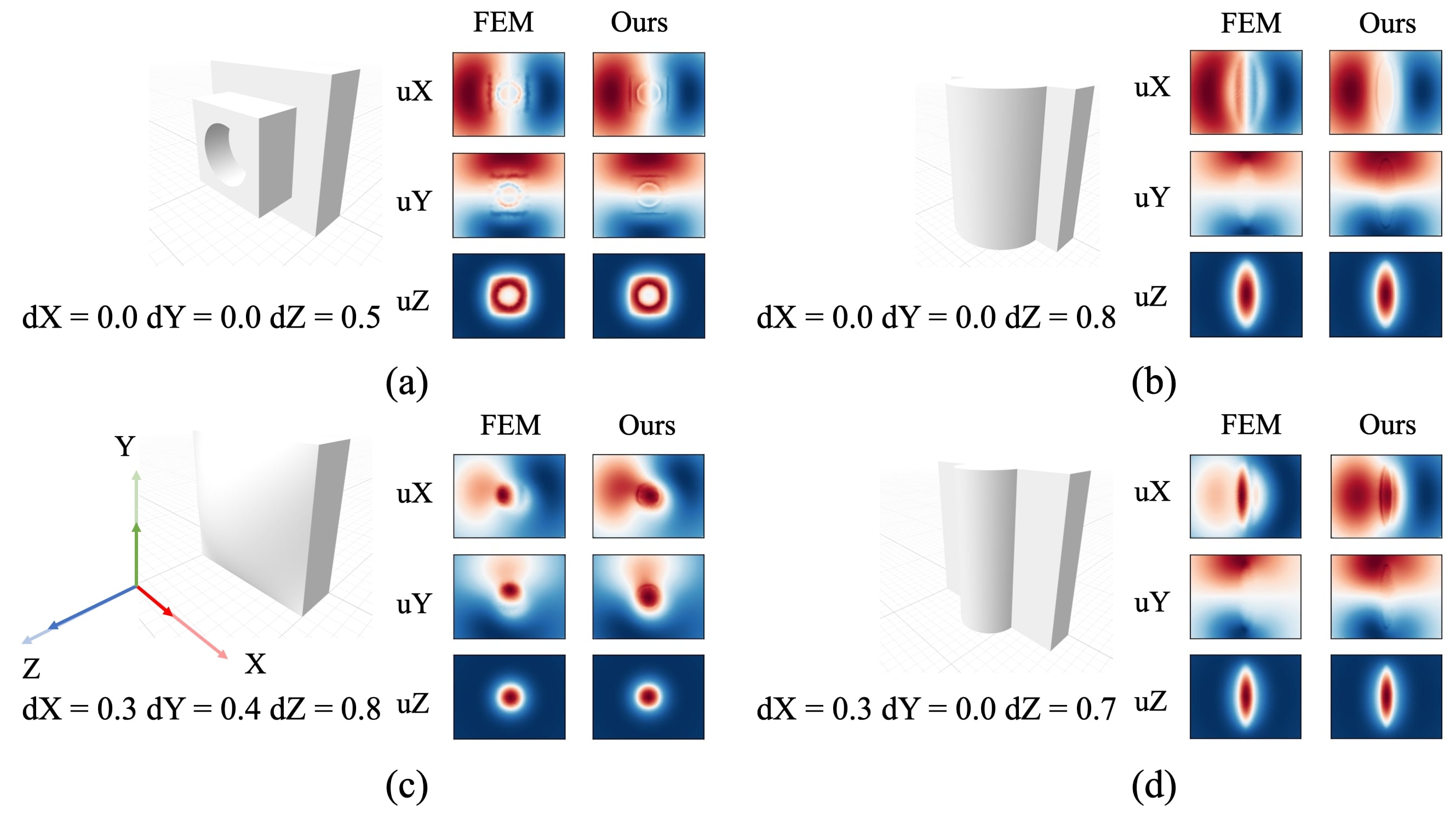}
    \caption{The simulated marker motion field in a dense mesh in comparison with the FEM data. The heat maps show the marker motion field in X, Y and Z directions where positive Z is the direction of normal loads and X, Y are the direction of shear loads.}
    \label{fig:dense}
    \vspace{-2mm}
\end{figure}

\subsection{Marker Motion Field Simulation}
We evaluate the simulation results with two references: 1) the dense displacement map generated by the FEM simulation, and 2) the sparse displacement map collected from a real sensor.
The contact cases are with objects in Fig.~\ref{fig:dataset} under combinations of different normal loads and shear loads. The load displacement varies from 0.3 mm to 0.8 mm.

\textbf{Comparison with FEM simulation}
As illustrated in the four sets of comparison from Fig.~\ref{fig:dense}, the dense mesh vertices displacements on X, Y, Z are simulated from both the FEM (Fig.~\ref{fig:dense} (b) left) and our methods (Fig.~\ref{fig:dense} (b) right). The color red means negative displacement value and blue means positive values. The average interpolated pixel-wised L1 errors over the gelpad surface on dataset are $3.58\times 10^{-3}$ mm for X-axis, $3.32\times 10^{-3}$ mm for Y-axis, $5.43\times 10^{-3}$ mm for Z-axis, and $5.40\times 10^{-3}$ mm for XY (gelpad surface). 

\textbf{Comparison with both real data and FEM simulation}
Examples of results are shown in Fig.~\ref{fig:all_markers_1}. The mean of marker motion's magnitude L1 errors on dataset is $1.00\times 10^{-2}$ mm between real \& FEM, $1.02\times 10^{-2}$ mm between real \& ours, and $3.96\times 10^{-3}$ mm between FEM \& ours. We weight the marker motion's angular errors based on the its magnitude because smaller marker motion is easier being affected by the system noise. The weighted mean of marker motion's angular L1 errors is $12.94^{\circ}$ between real \& FEM, $14.57^{\circ}$ between real \& ours, and $4.89^{\circ}$ between FEM \& ours. From the experimental results, the FEM model and our model match well, but there is still a gap from the simulation to the real gelpad soft body model. Three reasons observed from the experiments causing the errors are: 1) The gelpad is hand-manufactured and it is not perfectly matched with the FEM model in ANSYS. 2) The marker motions tracked from the real sensor's data have the noise in marker extraction and tracking. 3) When shear loads are present, our model cannot model the partial slip but it is very common for the real contact cases.

\textbf{Speed Testing} Our dense marker motion field simulation runs 9.22 seconds on average tested with CPU only. {The FEM simulation in ANSYS with CPU costs 2 to 4 hrs for difference cases.} In addition, according to Narang \etal\cite{narang2021sim}'s 5.57 seconds per sim for BioTac sensor in Isaac Gym with GPU acceleration, our simulation has a reasonable low computing demand.

We show some final results that combine the optical simulation and marker motion simulation in Fig.~\ref{fig:all_markers_1}.

\section{Conclusion}
We present Taxim, an example-based GelSight tactile simulation model that combines optical and marker motion field simulation. We construct a polynomial table to simulate the optical response of GelSight from the contact geometry, and apply the linear displacement relationship and the superposition principle to simulate the markers' motion. Our simulation is computationally light weight, easy to set up and use, and simple to apply to different sensors. It also incorporates the sensor's illumination features and system noise through calibration with examples from real sensors. We have shown that our optical simulation outperforms the other state-of-the-art tactile simulations, and our marker motion field simulation achieves high accuracy by evaluating on a self-designed dataset. To the best of our knowledge, this is the first integrated work considering both the optical and marker motion simulation. 

{To extend this work, we plan to apply different sim-to-real robot perception and manipulation tasks using our simulation model. In addition, currently we simulate the quasi-static contact, and we plan to investigate the dynamic contact process and simulate the dynamic phenomena such as slip. The simulation pipeline can be computationally improved by applying GPU acceleration.}

\addtolength{\textheight}{-0cm}   




\section*{ACKNOWLEDGMENT}
This work is supported by Facebook. The authors sincerely thank Arpit Agarwal, Tim Man, Yufan Zhang, Xiaofeng Guo, Uksang Yoo, Raj Kolamuri, Ruijia Xing, William Yan, Alankar Kotwal, Ioannis Gkioulekas, Mononito Goswami, Sudharshan Suresh, Haomin Shi for all the help on the discussion, experiment setup and manuscript revision.


\bibliographystyle{bib/IEEEtran}
\bibliography{bib/IEEEabrv, bib/ref}

\end{document}